\documentclass[letterpaper]{article} 
\usepackage{aaai24}  
\usepackage{times}  
\usepackage{helvet}  
\usepackage{courier}  
\usepackage[hyphens]{url}  
\usepackage{graphicx} 
\usepackage{natbib}  
\usepackage{caption} 
\usepackage{algorithm}
\usepackage{algorithmic}

\usepackage{newfloat}
\usepackage{listings}
\DeclareCaptionStyle{ruled}{labelfont=normalfont,labelsep=colon,strut=off} 
\floatstyle{ruled}
\newfloat{listing}{tb}{lst}{}
\floatname{listing}{Listing}
\usepackage{amsfonts}
\usepackage{amsmath}
\usepackage{algorithm}
\usepackage{algorithmic}
\usepackage{graphicx}
\usepackage{subcaption}
\usepackage{multirow}
\usepackage{array}
\usepackage{color, colortbl}
\usepackage{booktabs}

\definecolor{ashgray}{rgb}{0.7,0.75,0.71}
\definecolor{babypink}{rgb}{0.96,0.76,0.76}
\definecolor{taupegray}{rgb}{0.55, 0.52, 0.54}
\definecolor{gainsboro}{rgb}{0.86, 0.86, 0.86}
\definecolor{lightgray}{rgb}{0.83, 0.83, 0.83}
\definecolor{indigo(dye)}{rgb}{0.0, 0.25, 0.42}
\definecolor{silver}{rgb}{0.75, 0.75, 0.75}

\usepackage[final]{changes}

\usepackage{todonotes}

\title{LgTS: Dynamic Task Sampling using LLM-generated sub-goals  for Reinforcement Learning Agents}
\author{
    \hspace{1em} Yash Shukla\textsuperscript{\rm 1} \hspace{3em}
    Wenchang Gao\textsuperscript{\rm 1} \hspace{3em}
    Vasanth Sarathy\textsuperscript{\rm 1}  \hspace{12em}
    Alvaro Velasquez\textsuperscript{\rm 2} \hspace{2em}
    Robert Wright\textsuperscript{\rm 3} \hspace{3em}
    Jivko Sinapov\textsuperscript{\rm 1}
}
\affiliations{
\vspace{0.5em}
    \textsuperscript{\rm 1}Tufts University \hspace{3em}
    \textsuperscript{\rm 2}University of Colorado Boulder \hspace{3em}
    \textsuperscript{\rm 3}Georgia Tech Research Institute \hspace{3em}

}

\usepackage{bibentry}

\DeclareMathOperator*{\argmax}{arg\,max}

\begin{document}

\maketitle 
\begin{abstract}
Recent advancements in reasoning abilities of Large Language Models (LLM) has promoted their usage in problems that require high-level planning for robots and artificial agents. However, current techniques that utilize LLMs for such planning tasks make certain key assumptions such as, access to datasets that permit finetuning, meticulously engineered prompts that only provide relevant and essential information to the LLM, and most importantly, a deterministic approach to allow execution of the LLM responses either in the form of existing policies or plan operators. In this work, we propose LgTS (LLM-guided Teacher-Student learning), a novel approach that explores the planning abilities of LLMs to provide a graphical representation of the sub-goals to a reinforcement learning (RL) agent that does not have access to the transition dynamics of the environment. The RL agent uses Teacher-Student learning algorithm to learn a set of successful policies for reaching the goal state from the start state while simultaneously minimizing the number of environmental interactions. Unlike previous methods that utilize LLMs, our approach does not assume access to a propreitary or a fine-tuned LLM, nor does it require pre-trained policies that achieve the sub-goals proposed by the LLM. Through experiments on a gridworld based DoorKey domain and a search-and-rescue inspired domain, we show that generating a graphical structure of sub-goals helps in learning policies for the LLM proposed sub-goals and the Teacher-Student learning algorithm minimizes the number of environment interactions when the transition dynamics are unknown.    
\end{abstract}





         








\section{Introduction}

Large Language Models (LLMs) have been trained on a large corpus of natural language data that enables them to reason, engage in dialogue and answer questions based on a user specified prompt~\cite{touvron2023llama, zhao2023survey}. Recently, several techniques have utilized such LLMs to command artificial agents to perform a set of household tasks using natural language~\cite{ahn2022can,singh2023progprompt,yang2023llm}. To make such LLM-guided task solving techniques work, current approaches utilize several engineered tools (e.g. prompt engineering~\cite{wei2022chain}, LLM finetuning~\cite{peng2023instruction}) that aid the LLM-guided agent to perform anticipated actions~\cite{singh2023progprompt, driess2023palm}. First, language guided agents require a large number of labeled examples that associate a natural language prompt to a set of trajectories or to a successful policy that satisfies the natural language prompt. This costly procedure requires the human engineer to collect successful trajectories for each natural language label. To prevent the LLM from proposing unreasonable responses/plans, these techniques require fine-tuning the LLM on a labeled dataset of prompt-response pairs~\cite{song2022llm}. Moreover, current approaches
query the LLM to produce a single static plan (e.g.,~\cite{ding2023task,ahn2022can}) from the language instruction and assume that the agent is capable of executing the plan. This practice does not consider the environmental configuration, and the same language instruction can produce plans that are sub-optimal for different configurations. LLM-guided approaches used for task planning assume access to the Planning Domain Definition Language (PDDL)~\cite{liu2023llm+} that informs the LLM about which high-level actions (operators) are available, what is the cost of following a particular path etc. In absence of high level operators, iteratively querying the LLM once a RL policy fails to satisfy a sub-goal is excessively expensive.

\begin{figure*}[t]
	\centering
	\begin{minipage}{.25\textwidth}
		\includegraphics[width=\textwidth,height=0.9\textwidth]{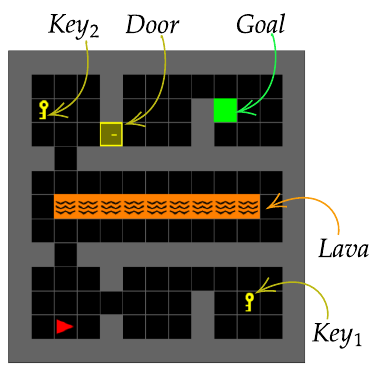}
        \subcaption{Gridworld domain}
        \label{fig:minigrid}
	\end{minipage}%
	\centering
	\begin{minipage}{.7\textwidth}
		\centering
		\includegraphics[width=\textwidth,height=0.3\textwidth]{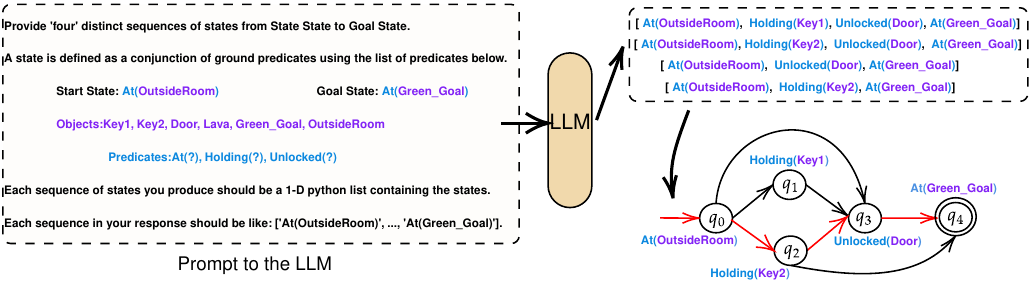}
        \vspace{0.1em}
        \subcaption{Prompt to the LLM (left) and the LLM output and corresponding DAG (right)}
        \label{fig:minigridprompt}
	\end{minipage}%
	\hfill
	\centering
	\caption{(a) Gridworld domain and descriptors. The agent (red triangle) needs to collect one of the keys and open the door to reach the goal; (b) The prompt to the LLM that contains information about the number of paths $n$ expected from the LLM and the symbolic information such as the entities, predicates and the high-level initial and goal states of the of the environment (no assumptions if the truth values of certain predicates are unknown). The prompt from the LLM is a set of paths in the form of ordered lists. The paths are converted in the form of a DAG. The path chosen by \emph{LgTS} is highlighted in red in the DAG in Fig.~\ref{fig:minigridprompt}} 
 \label{fig:overview}
\end{figure*}

\begin{figure}[t]
	\centering
		\includegraphics[width=0.45\textwidth,height=0.27\textwidth]{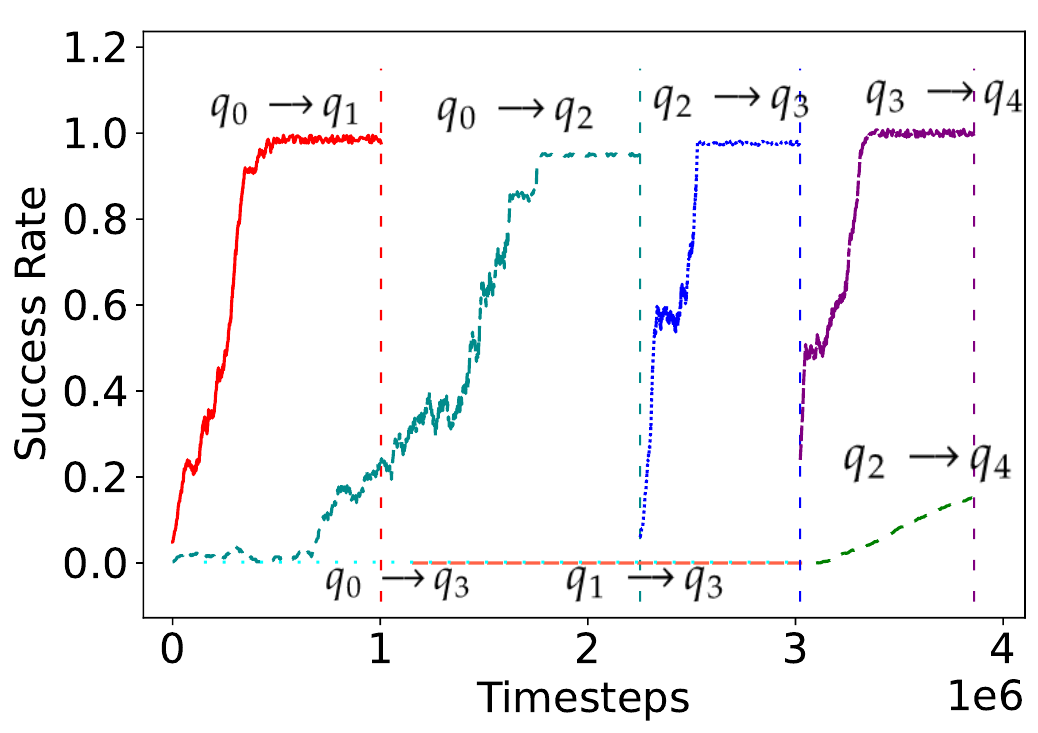}\label{fig:minigrid-ltl}
	\caption{Learning curves for individual
sub-tasks (in different colors) generated using \emph{LgTS}} 
 \label{fig:overview-3}
\end{figure}

In our work, we ease the above mentioned limitations by querying an off-the-shelf LLM to produce multiple feasible paths that have the potential to lead to the goal state. Introducing redundancy by querying multiple paths helps the agent explore several sub-goals in the environment and use that knowledge to figure out which sub-goal sequence will satisfy the goal objective. These multiple paths can be represented using a directed acyclic graph where the nodes of the graph are the sub-goal states proposed by the LLM and an edge of the graph is considered as a sub-task. A trajectory induced by a successful policy for the sub-task transitions the agent form one high-level state to another. The RL agent aims to learn a set of policies for one of the paths proposed by the LLM. Since the RL agent does not have information about the transition dynamics of the environment, the paths proposed by the LLM do not convey information about which path is more feasible and the RL agent needs to explore the environment to find out a feasible path. The paths proposed by the LLM can be sub-optimal and hence it is essential to minimize the number of times the agent interacts with the environment. Minimizing the overall number of environmental interactions while learning a set of successful policies is non-trivial as this problem is equivalent to finding the shortest path in a graph where the edge weights are unknown a priori~\cite{szepesvari2004shortest}. In our case, the edge weight denotes the total number of environmental interactions required by the agent to learn
a successful policy for an edge in the graph, in which the agent must
induce a visit to a state where certain properties hold true.
Additionally, we can only sample interactions for a sub-task
if we have a policy that can reach the edge’s source node
from the start node of the graph. 

The high-level overview of our approach is given in Fig.~\ref{fig:overview}. As an example, let us look at the environment in Fig.~\ref{fig:minigrid}. The goal for the agent is to collect \emph{any} of the two \emph{Keys}, followed by opening the \emph{Door} and then reaching the \emph{Goal} while avoiding the \emph{Lava} at all times. Our prompt to the LLM (Fig.~\ref{fig:minigridprompt}) contains information about the entities that are present in the environment along with the predicates the agent can identify. We assume that the agent has access to sensors that can detect entities in the environment and can also determine whether a certain predicate is true or not. The LLM is not provided the type hierarchy i.e., it does not have the information that associates predicates to its applicable entities. We assume access to a labeling function that maps an environmental state to a high-level symbolic state that informs the agent which predicates are true. The LLM outputs a number of ordered sub-goal sequences (lists) that can potentially be the paths of satisfying the high level goal. Each element in the list (a high-level state) is a conjunction of predicates and entities and satisfying this high-level state is a sub-goal for the RL agent (Fig.~\ref{fig:minigridprompt}). These set of sequences can effectively be converted to a directed acyclic graph where the start node of the graph is the initial high-level state of the agent and the goal node is the final high-level state (Fig.~\ref{fig:minigridprompt}). Each path in the graph is a sequence of states proposed by the LLM. 

The RL agent interacts with the environment to find a set of successful policies that guide the agent from the high-level start state to the high-level goal state. The LLM is not provided with information about the environment configuration, such as: the optimal number of interactions required to reach $Door$ from $Key_1$ are much higher compared to the interactions required to reach $Door$ from $Key_2$, making the $Key_1$ to $Door$ trajectory sub-optimal. Hence, while interacting with the environment, it is crucial to prevent any additional interactions the agent spends in learning policies for sub-tasks (individual edges in the DAG) that ultimately do not contribute to the final path the agent takes. That is, the agent should realize that the individual transitions $q_1 \rightarrow q_3$, $q_0 \rightarrow q_3$ and $q_2 \rightarrow q_4$ will take longer to train, and hence the amount of time spent in learning them should be minimized.

 To tackle this problem, we employ an adaptive Teacher-Student learning strategy, where, (1) the Teacher agent uses its high-level policy along with exploration techniques to actively sample a sub-task for the Student agent to learn. The high-level policy considers the graphical representation and the Student agent's expected performance on all the sub-tasks, and aims to satisfy the high-level objective in the fewest number of interactions, and (2) the Student agent interacts with the environment for a few steps (much fewer than the interactions required to learn a successful policy for the sub-task) while updating its low-level RL policy for the sampled sub-task. The Teacher observes the Student's performance on these interactions and updates its high-level policy. Steps (1) and (2) continue alternately until the Student agent learns a set of successful policies that guide the agent to reach an environmental goal state. The trajectory given by a successful RL policy for a sub-task (an edge) switches the agent's high-level state from the edge's source node to the edge's destination node.

 Our proposed approach, \emph{LgTS} begins with the aim of learning three distinct policies $\pi_{01}$ for the task of visiting $q_1$ from $q_0$, $\pi_{02}$ for the task of visiting $q_2$ from $q_0$, $\pi_{03}$ for the task of visiting $q_3$ from $q_0$, avoiding $Lava$ at all times. The Teacher initially samples evenly from these three sub-tasks but later biases its sampling toward the sub-task on which the Student agent shows higher learning potential. Once the agent learns a successful policy for one of the sub-tasks (let's say the learned policy $\pi_{01}^*$ satisfies the transition $ q_0 \rightarrow q_1$), the Teacher does not sample that task anymore, identifies the next task(s) in the graphical representation, and appends them to the set of tasks it is currently sampling (in this case, the only next task is: $q_1 \rightarrow q_3$). Since the agent only has access to the state distribution over $q_0$, it follows the trajectory given by $\pi_{01}^*$ to reach a state that lies in the set of states where $q_1$ holds true before commencing its learning for the policy ($\pi_{13}$) for $q_1 \rightarrow q_3$. If the agent learns the policies $\pi_{02}^*$ for satisfying the sub-task defined by $ q_0 \rightarrow q_2$ and $\pi_{23}^*$ for $ q_2 \rightarrow q_3$ before learning $\pi_{13}$ and $\pi_{03}$, it effectively has a set of policies to reach the node $q_3$. Thus, the Teacher will now only sample the next task in the graphical representation $q_3 \rightarrow q_4$, as learning policies for paths that reach $q_3$ are effectively redundant. This process continues iteratively until the agent learns a set of policies that reach the goal node ($q_4$) from the start node ($q_0$). The learning curves in Fig.~\ref{fig:overview-3} empirically validate the running example. The agent learns policies for the path $q_0 \rightarrow q_2 \rightarrow q_3 \rightarrow q_4$ that produce trajectories to reach the goal node $q_4$ from the initial node $q_0$, without excessively wasting interactions on the unpromising sub-tasks $q_1 \rightarrow q_3$, $q_0 \rightarrow q_3$ and $q_2 \rightarrow q_4$. The dashed lines in Fig.~\ref{fig:overview-3} signify the interactions at which a task policy converged.

In this work, we propose LgTS, a framework for generating a graphical representation of sub-tasks through an LLM response and then using this graph to learn policies to satisfy the goal objective. Through experiments, we demonstrate that LgTS is able to construct sub-tasks using the LLM output, realize which of these sub-tasks are unpromising, and only learn policies that will satisfy the goal while minimizing the number of environmental interactions. 

\section{Related Work}

Large Language Models (LLMs)~\cite{min2021recent} are trained on a huge corpora of natural language prompts that enables them to answer questions, reason and engage in dialogue. Being trained on vasts amounts of data enables LLMs to be applicable in tasks that require a general understanding of the task and the surrounding. Recently, several approaches have used LLMs for robot planning tasks where the role of LLM is to decompose the natural language instruction into language that can directly be fed to an artificial agent or a robot for execution~\cite{brohan2023rt,ahn2022can,ding2023leveraging}. Applications of LLMs in embodied agents include error correction during long horizon planning~\cite{raman2022planning}, intrinsically shaping the reward to promote exploration~\cite{kwon2023reward}, object rearrangement in robotics~\cite{stone2023open} and for augmenting a symbolic planner by providing a workable symbolic plan~\cite{liu2023llm+}. These techniques rely on several strongly engineered tools such as chain-of-thought prompting, fine-tuning of datasets and assumption of a verifier (such as a symbolic planner) that can determine if an LLM-generated plan can succeed~\cite{song2022llm}. Absence of these techniques significantly reduces the accuracy of the LLM-generated plans. If we do not have access to the high-level operators/actions that can solve the plan, it is very difficult to verify the correctness of the plan. Instead of completely relying on a single plan generated by the LLM, our approach queries multiple plans from an off-the-shelf LLM (no finetuning) that have the potential to satisfy the high-level goal objective. We then construct a graphical representation from the LLM output and use an adaptive Teacher-Student learning algorithm to learn a set of policies that can satisfy the goal objective.

Another line of research has investigated representing the goal using high-level specification languages, such as finite-trace Linear Temporal Logic (LTL$_f$)~\citep{de2013linear} or Reward Machines (RM)~\citep{icarte2022reward,toro2018teaching,bozkurt2020control, xu2019transfer,alur2022framework,de2019foundations} that allow defining the goal using a graphical representation informing the agent which sub-goals must be achieved prior to attempting the final goal. Automaton-based RL approaches assume that the high-level goal is known before commencing the task, and that the goal objective can be represented using a set of formal language rules that build on sub-goals. Automaton-guided RL has been used for robotic domains~\citep{belta,li2017reinforcement} and for multi-agent settings~\citep{hammond2021multi}. RM approaches still require human guidance in defining the reward structure of the machine, which is dependent on knowing how much reward should be assigned for each sub-goal. Unlike the approaches mentioned above, our approach does not require a predetermined reward structure, nor does it assume that the graphical structure for the sub-goals is available in advance. Our method queries the LLM to provide us with multiple paths of reaching the final goal which we use to construct a graphical structure of sub-goals.

Teacher-Student algorithms~\citep{DBLP:journals/tnn/MatiisenOCS20} have been studied in Curriculum Learning literature~\citep{narvekar2020curriculum,shukla2022acute} and in the Intrinsic Motivation literature~\citep{oudeyer2009intrinsic}. The idea is to have the Teacher propose those tasks to Student on which the Student shows most promise. This strategy helps Student learn simpler tasks first, transferring its knowledge to complex tasks. The technique reduces the overall number of interactions necessary to learn a successful policy. These approaches tend to optimize a curriculum to learn a single policy, which does not scale well to temporally-extended tasks. Instead, we propose an LLM-guided Teacher-Student learning strategy that learns a set of policies for promising transitions in the sub-goal graph, promoting sample-efficient training compared to the baselines.

\section{Preliminaries}\label{sec:prelimiaries}

\textbf{Symbolic Planning:}
We assume access to symbolic information defined as ~$\Sigma=\langle\mathcal{E}, \mathcal{F}, \mathcal{Q}, q_0, q_g\rangle$, where $\mathcal{E}=\left\{\varepsilon_{1}, \ldots \varepsilon_{|\mathcal{E}|}\right\}$ is a finite set of known entities within the environment, and $\mathcal{F}=\left\{f_{1}(\odot), \ldots f_{|\mathcal{F}|}(\odot)\right\}$, where $ \odot \subset \mathcal{E}$, is a finite set of known predicates with their negations. Each predicate $f_{i}\left(\odot\right)$, along with its negation $\neg f_{i}(\odot)$, is contained in $\mathcal{F}$. $\mathcal{Q}=\left\{q_{1} \ldots q_{|\mathcal{Q}|}\right\}$ is the set of symbolic states in the environment. A symbolic state $q \in \mathcal{Q}$ is defined as a complete assignment over all predicate-entity pairs, i.e. $q = \bigcup^{|\mathcal{E}|}_{i = 1} \bigcup^{|\mathcal{F}|}_{j = 1} f_j(e_i)$. The starting state is given by $q_{0} \subset \mathcal{Q}$,  and,  $q_{g} \subset \mathcal{Q}$ is the goal state. We assume access to a \textit{single word} natural language description of the predicates and entities.
\\
\textbf{LLM prompt:}
Autoregressive LLMs are trained with
a maximum likelihood loss to model the probability of a
sequence of tokens $y$ conditioned on an input sequence $x$, s.t. $\phi = \argmax_{\phi}P(y | x, \phi)$, where $\phi$ are the LLM parameters. We assume access to a prompt function
$f_{p}: (\Sigma, n) \rightarrow x$ that takes in the symbolic information $\Sigma$ along with the number of paths that the LLM should generate $n$ and produces a natural language prompt $x$  that serves as an input to the LLM. The prompt to the LLM $x$ is designed in such a way that the output from the LLM $y = \mathsf{LLM}_{\phi} (x)$ can be construed into a graph. The LLM output $y$ is converted to a set of ordered lists, where each element of the list is a high-level state $q \in \mathcal{Q}$. The first element of the list is the start state $q_0$ and the final element of the list is the goal state $q_g$. If any of the lists in the output $y$ do not satisfy the start state and the goal state conditions, i.e. if the first element of the LLM-generated list is not $q_0$ and the final element is not $q_g$, the LLM is reprompted to produce another ordered list as a response where the first element is $q_0$ and the final element is $q_g$.

Additionally, we assume access to a function $\mathsf{Graph}: \mathsf{List}(y) \rightarrow \mathcal{U}$ that takes in the ordered lists generated from the LLM output $y$ as its input and produces a directed acyclic graph $\mathcal{U} = (V, E)$ where each vertex $v\in V$ is a high-level state, i.e., $V \subseteq \mathcal{Q}$, and the set of edges $E \subseteq V \times V$ connects two high-level states, i.e. $e_{ij}$ is a directed edge from $v_i$ to $v_j$ (see Fig.~\ref{fig:minigridprompt}). 
\\
\textbf{Labeled MDP:}
An episodic labeled Markov Decision Process (MDP) $M$ is a tuple $(\mathcal{S}, \mathcal{A}, \tau, R, \mathcal{S}_0, \mathcal{S}_f, \gamma, T, \mathcal{Q}, L)$, where $\mathcal{S}$ is the set of MDP (low-level) states, $\mathcal{A}$ is the set of actions, $p(s'|s,a)$ is the transition probability of reaching state $s' \in \mathcal{S}$ from $s \in \mathcal{S}$ using action $a \in \mathcal{A}$, $R: \mathcal{S} \times A \times \mathcal{S} \rightarrow \mathbb{R}$ is the reward function, $\mathcal{S}_0$ are the initial states, $\mathcal{S}_f$ are the terminal states, $\gamma \in  [0,1]$ is the discount factor, $T$ is the maximum number of interactions in any episode, $\mathcal{Q}$ is the set of high-level states, and $L: \mathcal{S} \rightarrow \mathcal{Q}$ is a labeling function that maps an MDP state $s \in \mathcal{S}$ to a high-level state $q \in \mathcal{Q}$.

In every interaction, the agent observes the current state $s$ and selects an action $a$ according to its policy function $\pi(a|s, \theta)$ with parameters $\theta$. The MDP transitions to a new state $s' \in \mathcal{S}$ with probability $\tau(s' \mid s, a)$. The agent's goal is to learn an \emph{optimal policy $\pi^*$} that maximizes the discounted return $G_0 = \sum^{T}_{t = 0}\!\gamma^t\! R(s'_t,a_t,s_t) $ until the end of the episode, which occurs after at-most $T$ interactions.

\subsection{Problem Statement}

Given the symbolic information $\Sigma$ and access to large-language model $\mathsf{LLM}_{\phi}$, the aim of this work is to:
\begin{itemize}
    \item Convert the LLM output $y = \mathsf{LLM}_{\phi}(f_p (\Sigma, n))$ into a directed acyclic graph $\mathcal{U} = (V,E)$ such that each vertex $v \in V$ is a high-level state $q\in \mathcal{Q}$ and $E \subseteq V \times V$.
    \item Construct a sub-task MDP $\mathsf{Task}(v_i,v_j)$ corresponding to each transition of the graph. A sub-task defined by an edge $e_{ij}$ from node $v_i$ to $v_j \in V$ defines a reach-avoid objective for the agent represented by the formula:
    \begin{equation}\label{eq1}
        \mathsf{Task}(v_i, v_j) = \mathsf{\textbf{F}}(v_j) \land \mathsf{\textbf{G}}\left(\bigwedge\limits_{r \in \mathsf{Succ}(v_i), r \neq v_j} \neg v_r\right)        
    \end{equation}

    where $v_j$ is the symbolic state corresponding to the destination node of edge $e_{ij}$ and $\mathsf{Succ}(v_i)$ is the set of states of successors of node $v_i$ in the DAG $\mathcal{U}$, \textbf{F} and \textbf{G} correspond to \textit{Eventually} and \textit{Always} respectively. The $\mathsf{Task}(v_i,v_j)$ represents an MDP $M'$ where all initial states $s' \in \mathcal{S}^{M'}_0$ are mapped to the high-level state $L(s') \rightarrow v_i$ and terminal states $s' \in \mathcal{S}^{M'}_f$ of $\mathsf{Task}(v_i,v_j)$ are mapped to the high-level state $L(s') \rightarrow v_j$.
    $\mathsf{Task}(e_{ij})$ and $\mathsf{Task}(v_i, v_j)$ are used interchangeably.
    
    
    \item Learn a set of policies $\pi_{ij}^*$, $i,j=0, \ldots, n-1$, with the following three properties:\\ 
        (i) Following $\pi_{01}^*$ results in a trajectory in the task MDP $\mathsf{Task}(e_{01})$ that induces a transition from $q_0$ to some state $q_1 \in Q$ in the DAG, following $\pi_{12}^*$ results in a path in MDP $\mathsf{Task}(e_{12})$ that induces a transition from $q_1$ to some state $q_2 \in Q$ in the DAG, and so on. (ii) The resulting path $q_0 q_1 \ldots q_g$ in the DAG terminates at the goal state $q_g$, with probability greater than a given threshold, $\eta \in (0, 1)$. (iii) The total number of environmental interactions spent in exploring the environment and learning sub-task policies are minimized.      

\end{itemize}

\section{Methodology}

First, given the symbolic information $\Sigma = \langle\mathcal{E}, \mathcal{F}, \mathcal{Q}, q_0, q_g\rangle$ and $n$, the number of ordered lists of sub-goal paths we expect from the LLM, we generate a natural language prompt $x = f_p(\Sigma, n)$. An example of a prompt is shown in Fig.~\ref{fig:overview}. The prompt directs the LLM to produce outputs that is converted to a set of $n$ ordered lists, where each element in the list is a high-level state, the first element of the list is initial high-level state $q_0$, and the final element of the list is the goal high-level state $q_g$. 
This prompt is fed to a large language model (LLM) to produce a sequence of tokens where each token is given by $y = \mathsf{LLM}_{\phi}(f_p(\Sigma,n))$. For our work, we used LLAMA2~\cite{touvron2023llama}, an open-source LLM that allows version control and is easily accessible. While the output generated by the LLM depends on its training protocol and on the dataset used for its training, this work does not involve investigating and comparing the output from different LLMs as that is tangential to our study. 

The next step is to convert the natural language output from the LLM $y$ into a directed acyclic graph $\mathcal{U} = (V,E)$ such that each element $v \in V$ is a high-level state $q\in \mathcal{Q}$ and $E \subseteq V \times V$. If the output $y$ from the LLM does not satisfy the high-level initial and goal state conditions (see Sec.~\ref{sec:prelimiaries}), the LLM is reprompted until the output $y$ matches the correct syntax. We parse the output to get $n$ distinct paths of reaching the high-level goal state $q_g$ from the initial high-level state $q_0$. These $n$ distinct paths are in the form of an adjacency list for a graph. While constructing the graph, we omit self-loops and cycles, generating a directed acyclic graph (DAG) $\mathcal{U}$. 

Given the DAG $\mathcal{U}$, we define a set of sub-tasks based on the edges of the DAG. Intuitively, given the current MDP state for the agent $s \in \mathcal{S}$ and its corresponding DAG node $L(s) \rightarrow q$, a sub-task defined by an edge from node $q$ to $p \in Q$ defines a reach-avoid objective for the agent represented by the formula~\ref{eq1}.

Each sub-task $\mathsf{Task}(q, p)$ defines a problem to learn a policy $\pi_{(q, p)}^*$ such that, given any MDP state $s_0$ that maps to the high-level state $q$ (i.e., $L(s_0)\!\! \rightarrow\! q$), following $\pi_{(q, p)}^*$ results in a path $s_0 s_1 \ldots s_n$ in MDP that induces the symbolic path $q q \ldots q p$. That is, the high-level state of the agent remains at $q$ until it transitions to $p$. 



The algorithm for \emph{AGTS} is described in Algo.~\ref{alg:agts}. We begin by initializing the following quantities (lines 2-4):
    (1) Set of: Active Tasks $\mathcal{AT}$, Learned Tasks $\mathcal{LT}$, Discarded Tasks $\mathcal{DT}$;
(2) A Dictionary $P$ that maps a sub-task $\mathsf{Task}(e)$ of the DAG $\mathcal{U}$ to a policy $\pi_e$; (3) A Dictionary $Q$ that represents the Teacher Q-Values by mapping the learning progress (in terms of success rate) of the sub-task $\mathsf{Task}(e)$ to a numerical Q-value associated with that sub-task.  

Firstly, we convert $\mathcal{U}$ into an Adjacency Matrix $\mathcal{X}$ (line 6), and find the set of all the outgoing edges $E_{q_0} \subseteq E$ from the initial node $q_0$ (line 7). Satisfying the edge's formula $e_{q_0,q_1} \in E_{q_0}$ represents a reachability sub-task $M'$ where each goal state $s \in \mathcal{S}^{M'}_f$ of $M'$ satisfies the condition $L(s)\rightarrow q_1 $. The agent receives a positive reward for satisfying $\mathsf{Task}(e_{q_0,q_1})$ and a small negative reward at all other time steps. The state space, the action space and the transition dynamics of MDP $M'$ are equivalent to MDP $M$.  To complete the sub-task, the agent must learn a policy $\pi_{(q_0, q_1)}^*$ that ensures a visit from $q_0$ to $q_1$ with probability greater than a predetermined threshold ($\eta$). Moreover, the policy must also avoid triggering unintended transitions in the DAG. For instance, while picking up $Key_1$, the policy must not inadvertently pick up $Key_2$ as evident from the task objective formula~\ref{eq1}. 


We set the Teacher Q-Values for all the sub-tasks corresponding to edges in $\mathcal{AT}$ (i.e., $\forall e \in E_{q_0}$) to zero (line 8).
We formalize the Teacher's goal of choosing the most promising task as a \textit{Partially Observable MDP}~\citep{KAELBLING199899}, where the Teacher does not have access to the entire Student agent state but only to the Student agent's performance on a task (e.g., success rate or average returns), and as an action, chooses a task $\mathsf{Task}(e) \in$ $\mathcal{AT}$ the Student agent should train on next. In this POMDP setting, each Teacher action (a sub-task) has a Q-Value associated with it. Intuitively, higher Q-Values correspond to tasks on which the Student agent is more successful, and the Teacher should sample such tasks at a higher frequency to reach the goal node $q_g$ in fewest overall interactions.

(Step i) Given the Teacher Q-Values, we sample a sub-task $\mathsf{Task}(e) $\\$\in $ $\mathcal{AT}$ using the $\epsilon-$greedy exploration strategy (line 10), and (Step ii) The Student agent trains on task $\mathsf{Task}(e)$ using the policy $P[e]$ for `$x$' interactions (line 11). In one Teacher timestep, the Student trains for $x$ environmental interactions. 
Here, $x << $ total number of environmental interactions required by the agent to learn a successful policy for $\mathsf{Task}(e)$, since the aim is to keep switching to a task that shows highest promise. (Step iii) The Teacher observes the Student agent's average return $g_{t'}$ on these $x$ interactions, and updates its Q-Value for $\mathsf{Task}(e)$ (line 12):
\begin{equation}\label{eq2}
    Q[e] \leftarrow \alpha (g_{t'}) + (1-\alpha)Q[e]
\end{equation}
where $\alpha$ is the Teacher learning rate, $g_{t'}$ is the average Student agent return on $\mathsf{Task}(e)$ at the Teacher timestep $t'$. As the learning advances, $g_{t'}$ increases, and hence we use a constant parameter $\alpha$ to tackle the non-stationary problem of a moving return distribution. Several other algorithms could be used for the Teacher strategy (e.g., UCB~\citep{agrawal2012analysis}, Thomspson Sampling~\citep{auer2002finite}). Steps i, ii and iii continue successively until the policy for \textit{any} $\mathsf{Task}(e)\in $$\mathcal{AT}$ task converges.


We define a policy for $\mathsf{Task}(q,p)$ to be converged (line 13) if a trajectory $\omega$ produced by the policy triggers the transition with probability Pr$_{\omega \in \Omega}[\omega \mbox{ satisfies} \: \mathsf{Task}(q,p)]\:\geq \:\eta$ and $\Delta(g_{t'}g_{t'-1})\: < \:\mu$ where $\eta$ is the expected performance and $\mu$ is a small numerical value. Intuitively, a converged policy attains an average success rate $\geq \eta$ and has not improved further by maintaining constant average returns. Like other Reward Machine (RM) and automaton-based approaches, we assume access to the labeling function $L$ to examine if the trajectory $\omega$ satisfies the transition corresponding to the edge $e_{(q,p)}$ by checking if the final environmental state $s$ of the trajectory satisfies the condition $L(s)\! \rightarrow\! p$. The sub-goal regions need not be disjoint, i.e., the same state $s$ can satisfy predicates for multiple DAG nodes.
Once a policy for the $\mathsf{Task}(q,p)$ converges, we append $\mathsf{Task}(q,p)$ to the set of Learned Tasks $\mathcal{LT}$ and remove it from the set of Active Tasks $\mathcal{AT}$ (line 14). To ensure that the learned task does not get sampled any further, we set the Teacher Q-value for this sub-task to $-\infty$ (line 15). Once we have a successful policy for the $\mathsf{Task}(q,p)$ (the transition $q \rightarrow p$), we determine the sub-tasks that can be discarded (line 16). We find the sub-tasks corresponding to edges that: (1) lie before $p$ in a path from $q_0$ to the goal state $q_g$, and, (2) do not lie in a path to $q \in \mathcal{Q}$ that does not contain $p$. Intuitively, if we already have a set of policies that can generate a successful trajectory to reach the node $p$, we do not need to learn policies for sub-tasks that ultimately lead \textit{only} to $p$. We add all such sub-tasks to the set of Discarded Tasks $\mathcal{DT}$ (line 17), and set the Teacher Q-values for all the discarded tasks to $-\infty$ to prevent them from being sampled for the Student learning agent (line 18). 

By discarding such sub-tasks, we might fail to explore certain sub-tasks which could have led to an optimal or near-optimal path from the start to the goal node. In this work, our aim is not to find optimal policies but to learn policies that reach the goal node with two important criterion: (1) The probability of generating a trajectory that reaches $q_g$ from $q_0$ is $\geq \eta$ and (2) The overall number of environmental interactions are minimized. 

\begin{algorithm}[t]
\caption{ \emph{LgTS} ( $\mathcal{U}, M, \eta, x$ )}
\label{alg:agts}
\raggedright \textbf{Output}: Set of learned policies : $\Pi^*$, Edge-Policy Dictionary $P$\\
\begin{algorithmic}[1] 
\STATE \textbf{Placeholder Initialization}:\\
\STATE Sets of: Active Tasks ($\mathcal{AT}$) $\leftarrow \emptyset$; \\Learned Tasks ($\mathcal{LT}$) $\leftarrow \emptyset$; Discarded Tasks ($\mathcal{DT}$) $\leftarrow \emptyset$
\\
\STATE Edge-Policy Dictionary $P :  \mathsf{Task}(e) \rightarrow \pi$
\STATE Teacher Q-Value Dictionary: $Q : \mathsf{Task}(e) \rightarrow -\infty $
\STATE \textbf{Algorithm:}
\STATE $\mathcal{X} \leftarrow $ {\tt Adjacency\_Matrix} $(\mathcal{U})$ 
\STATE $\mathcal{AT}$ $\leftarrow $ $\mathcal{AT}$ $\cup$ $\{ \mathcal{X}[q_0]\}$ 
\STATE $\forall \:  \mathsf{Task}(e) \in  $ $\mathcal{AT}$   $: Q[e] = 0$  
\WHILE{True} 
\STATE $e \leftarrow $ {\tt Sample}$(Q)$ 
\STATE $P[e], g \leftarrow $ {\tt Learn}$(M, \mathcal{U}, e, x, P)$ 
\STATE {\tt Update\_Teacher}$(Q, e, g)$  
\IF{{\tt Convergence}($Q, e, g, \eta$)}
\STATE $\Pi^* \leftarrow \Pi^* \cup  P[e]$ ;  $\mathcal{LT}$ $\leftarrow$ $\mathcal{LT}$ $ \cup \{\mathsf{Task}(e)\}$ ;  \\$\mathcal{AT}$ $\leftarrow$ $\mathcal{AT}$ $ \setminus \{\mathsf{Task}(e)\}$
\STATE $Q[e]  = -\infty$ 
\STATE $E_{\mathcal{DT}} \leftarrow $ {\tt Discarded\_Tasks}$(\mathcal{X}, e)$
\STATE $\mathcal{DT}$ $\leftarrow$ $\mathcal{DT}$ $\cup \:  E_{\mathcal{DT}}$ 
\STATE $\forall\: \mathsf{Task}(\overline{e})\: \in\: E_{\mathcal{DT}} : \: Q[\overline{e}] = -\infty \:$ 
\STATE $E_{\mathcal{AT}} \leftarrow$ {\tt Next\_Tasks} $(\mathcal{X}, e,$ $\mathcal{DT}$) 
\IF{$|E_{\mathcal{AT}}| = 0$} 
\STATE {\tt break} 
\ENDIF
\STATE $\forall\: \mathsf{Task}(\overline{e})\: \in\: E_{\mathcal{AT}} : Q[\overline{e}] = 0 $  
\STATE $\mathcal{AT}$$ \leftarrow$  $\mathcal{AT}$ $\cup \: E_{\mathcal{AT}}$ 
\ENDIF
\ENDWHILE
\STATE \textbf{return} $\Pi^*, P$
\end{algorithmic}
\end{algorithm}

Subsequently, we determine the next set of tasks $E_{\mathcal{AT}}$ in the DAG to add to the $\mathcal{AT}$ set (line 19). This is calculated by identifying sub-tasks corresponding to all the outgoing edges from $p$. Since the edge $e_{q,p}$ corresponds to the transition $q \rightarrow p$, we have a successful policy that can produce a trajectory that ends in the high-level state $p$, and $E_{\mathcal{AT}}$ corresponds to $\mathcal{X}[p] \setminus $$\mathcal{DT}$ i.e., sub-tasks corresponding to all the outgoing edges from $p$ that do not lie in the $\mathcal{DT}$ set. 

Once we identify $E_{\mathcal{AT}}$, we set the Teacher Q-values for all $\mathsf{Task}(\overline{e}) \in E_{\mathcal{AT}}$ to $0$ so that the Teacher will sample these tasks (line 23). We consider an episodic setting where the episode starts from a state $s\in\mathcal{S}_0$ where the high-level state $q_0$ holds true, and if the current sampled sub-task is $\mathsf{Task}(p,r)$, the agent follows a trajectory using corresponding learned policies from $\Pi^*$ to reach a MDP state where the high-level state $p$ holds true, and then attempts learning a separate policy for $\mathsf{Task}(p,r)$.

The above steps (lines 9-26) go on iteratively until $E_{\mathcal{AT}}$ is an empty set, which indicates we have no further tasks to add to our sampling strategy, and we have reached the goal node $q_g$. Thus, we {\tt break} from the {\tt while} loop (line 21) and return the set of learned policies $\Pi^*$, and edge-policy dictionary $P$ (line 27). From $P$ and $\Pi^*$, we get an ordered list of policies $\Pi^*_{list} = 
[\pi_{(q_1,q_2)}, \pi_{(q_2,q_3)}, \ldots,$\\$ \pi_{(q_{k-1},q_k)}]$ 
such that sequentially following $\pi \in \Pi^*_{list}$ generates trajectories that reach the high-level goal state $q_g$.  



\section{Experiments}

We aim to answer the following questions: (Q1) Does \emph{LgTS} yield sample efficient learning compared to other baseline approaches?
(Q2) How does \emph{LgTS} perform when distractor objects are present in the environment that are not essential for satisfying the high-level goal? 
(Q3) Does \emph{LgTS} yield sample efficient learning even when the prompt to the LLM is modified by using synonyms for objects and predicates?
(Q4) How does \emph{LgTS} scale when the environment is complex and the optimal plan is longer than the DoorKey task?
(Q5) What are certain failure cases of \emph{LgTS}?
 
\subsection{LgTS - DoorKey Domain}~\label{sec:results-gridworld}

To answer Q1, we evaluated \emph{LgTS} on a Minigrid~\citep{minigrid} inspired domain. The environment configuration is shown in Fig.~\ref{fig:overview}. Essentially, the agent needs to collect \textit{any} of the two available \emph{Keys} before heading to the \emph{Door}. After \emph{toggling} the \emph{Door} open, the agent needs to visit the \emph{Green\_Goal}. At all times, the agent needs to avoid the \emph{Lava} object. We assume an episodic setting where an episode ends if the agent touches the \emph{Lava} object, reaches the \emph{Green\_Goal} or exhausts the number of allocated interactions.

This is a complex sequential decision making problem for a reinforcement learning agent as the agent needs to perform a series of correct actions to satisfy the high-level objective, which is to navigate to any of the two keys, pick a key and then unlock the door. Then, navigate to reach the green-goal state. In this environment, the agent has access to three navigation actions: \emph{move forward}, \emph{rotate left} and \emph{rotate right}. The agent can also perfom: \emph{pick-up} action, which adds a \emph{Key} to the agent's inventory if it is facing the \emph{Key}, \emph{drop} places the \emph{Key} in the next grid if \emph{Key} is present in the inventory, and, \emph{toggle} that toggles the \emph{Door} (closed $\leftrightarrow$ open) only if the agent is holding the \emph{Key}. The agent can hold only one \emph{Key} in its inventory. Hence, it needs to perform the \emph{drop} action to drop a key present in its inventory before picking up another key.
For this environment, we assume a fully-observable setting where the environmental state is a low-level encoding of the image. For each cell in the grid, the low-level encoding returns an integer that describes the item occupying the grid, along with additional information, if any (e.g., the \emph{Door} state can be open or closed). 

The prompt to the LLM contains information about the high-level start state $At(OutsideRoom)$, the high-level goal state \\ $At(Green\_Goal)$, the entities present in the environment - $Key_1, $ \\ $Key_2, Door, OutsideRoom, Green\_Goal, Lava$, the predicates that the agent can detect through its sensors - $Holding(?), At(?), $ \\ $ Unlocked(?)$, and a hyperpameter that defines the number of feasible high-level sequences given by the LLM $n$. 
We performed grid-search to find the value of $n$. For our experiments, $n=4$.


For the RL pipeline, we use PPO~\citep{DBLP:journals/corr/SchulmanWDRK17}, which works for discrete and continuous action spaces. We consider a standard actor-critic architecture with 2 convolutional layers followed by 2 fully connected layers. For \emph{LgTS}, the reward function is sparse. The agent gets a reward of $(1 - 0.9\frac{(interactions \:taken)}{(interactions \:allocated)})$ if it achieves the goal in the sub-task, and a reward of $0$ otherwise. For individual tasks, $interactions\:allocated = 100$.
The agent does not receive any negative rewards for hitting the $Lava$. 

\subsubsection{\textbf{Baselines:}}
We compare our \emph{LgTS} method against four baseline approaches and an oracle approach.\footnote{More details on baselines are given in Appendix }:
\begin{enumerate}
    \item Learning from scratch (LFS) where the goal for the agent is to learn a single policy that learns to satisfy the final high-level goal state using RL without any shaped reward.
    \item Teacher-student curriculum learning (TSCL) appraoch where the Teacher agent samples most promising task (based on average success rate) without the use of any graphical structure to guide the learning progress of the agent. The set of tasks is chosen by a human expert. In our experiments, the set of tasks included every feasible transition in the automaton description of the task.
    \item \textbf{(Oracle approach)} Automaton-guided Teacher-Student learning (AgTS) where the graphical structure of the sub-goal is generated using the finite-trace Linear Temporal Logic (LTL$_f$) formula given by an oracle. For this task, the LTL$_f$ formula is: $\varphi_f :=  \textbf{G} \neg Lava \wedge \textbf{F}((Key_1 | Key_2 ) \wedge \:\mbox{\textbf{F}}(Door\: \& \:\mbox{\textbf{F}}(Goal)))$ where $G$ and $F$ represent \textit{Always} and \textit{Eventually} respectively. We use the equivalent DFA representation of the above LTL$_f$ formula as the graphical representation, and perform the Teacher-Student learning approach outlined in section 4.
    \item Automaton-guided Reward Shaping (AGRS) where the DFA representation of the LTL$_f$ formula is used as a reward shaping mechanism to guide the agent toward the final high-level goal state. The reward given to the agent is proportional to the distance from the goal node.
    \item LLM-guided Reward Shaping (LgRS) where the graph generated from the $n$ high-level sub-goal sequences is used as a reward shaping mechanism to guide the agent toward the final high-level goal state. The reward given to the agent is proportional to the distance from the goal node.
\end{enumerate}

\begin{table}
    		\centering
	\begin{tabular}{>{\centering}p{1.1cm}>{\centering}p{2.6cm}>{\centering}p{2.4cm}p{0.05cm}}
\arrayrulecolor{taupegray}\toprule
\textbf{Approach} & \textbf{$\#$ Interactions} \\ (Mean $\pm$ SD) & \textbf{Success Rate }\\ (Mean $\pm$ SD) \tabularnewline
\midrule
\rowcolor{gainsboro}
\emph{LgTS}  & $(3.98 \pm 0.42)\times 10^6$ &\centering $0.92 \pm 0.03$\tabularnewline
\rowcolor{gainsboro}
\emph{AgTS}  & $(2.67 \pm 0.36)\times 10^6$ &\centering $0.94 \pm 0.02$\tabularnewline
\emph{LFS}  &  $5 \times 10^7$ &\centering $0 \pm 0$\tabularnewline
AgRS  & $5 \times 10^7$ &\centering $0.05 \pm 0.04$ \tabularnewline
LgRS  & $5 \times 10^7$ &\centering $0 \pm 0$ \tabularnewline
TSCL  & $5 \times 10^7$ &\centering $0 \pm 0$ \tabularnewline
\bottomrule
\end{tabular}
		\caption{Table comparing $\#$interactions $\&$ success rate for the DoorKey domain.}
		\label{table:gridworld}
\end{table}
The results in Table~\ref{table:gridworld} (averaged over 10 trials) show that \emph{LgTS} reaches a successful policy quicker compared to the learning from scratch, teacher-student curriculum learning, and LLM-guided reward shaping baseline approaches.
We observe that the number of environmental interactions taken by our proposed approach are comparable to the automaton-guided teacher student (AgTS) algorithm where the ground truth graph is in the form of an automaton, and the graph is provided by an oracle.
Several of the other baseline approaches such as LFS, TSCL, LgRS, AgRS fail to learn a successful policy for reaching the high-level goal state demonstrating that approaches that tend to learn a single policy for the entire objective are unable to satisfy the goal condition. Reward shaping fails as agent greedily favours reaching the high level state $q_1$ over $q_2$ and is unable to reach node $q_3$ from $q_1$. 

We evaluated the average graph edit distance (GED) between the graphs generated using the $LgTS$ and the AgTS approach. The GED is the number of edge/node changes needed to make two graphs isomorphic. We observed an average graph edit distance of $2.1\pm0.2$. This indicates that the graph generated by the oracle through AgTS, which has five nodes and five edges, can be converted to the graph generated by LgTS by performing $\sim2.1$ changes. 

Refer Appendix Section B for additional experiments and discussions on how LgTS performs when the number of sub-goal path $n$ varies w.r.t to the number of objects and predicates. To summarize, we observed high interaction cost and low success rate when $n$ was too low (1 or 2), denoting that the LLM fails to consider different paths for satisfying the goal and generates a path that does not succeed given an unknown environment configuration. 

\subsection{LgTS with distractor entities}

To answer Q2, we evaluated LgTS on a task environment that contains entities that do not affect the optimal path for reaching the high-level goal state. During each run, the environment contains $1-3$ instances of distractor objects that are modeled in the LLM prompt and in the environment dynamics. For our experiment, the distractor items are household kitchen items such as apple, plate, fruit etc. Since the optimal path or the task solution has not changed, the paths suggested by the LLM should ignore the distractor objects. 

\begin{table}
    		\centering
	\begin{tabular}{>{\centering}p{1.1cm}>{\centering}p{2.6cm}>{\centering}p{2.4cm}p{0.05cm}}
\arrayrulecolor{taupegray}\toprule
\textbf{Approach} & \textbf{$\#$ Interactions} \\ (Mean $\pm$ SD) & \textbf{Success Rate }\\ (Mean $\pm$ SD) \tabularnewline
\midrule
\rowcolor{gainsboro}
\emph{LgTS}  & $(4.64 \pm 1.72)\times 10^6$ &\centering $0.84 \pm 0.08$\tabularnewline
\rowcolor{gainsboro}
\emph{AgTS}  & $(3.21 \pm 0.57)\times 10^6$ &\centering $0.90 \pm 0.90 $\tabularnewline
LFS  &  $5 \times 10^7$ &\centering $0 \pm 0$\tabularnewline
AgRS  & $5 \times 10^7$ &\centering $0 \pm 0$ \tabularnewline
LgRS  & $5 \times 10^7$ &\centering $0 \pm 0$ \tabularnewline
TSCL  & $5 \times 10^7$ &\centering $0 \pm 0$ \tabularnewline
\bottomrule
\end{tabular}
		\caption{Table comparing $\#$interactions $\&$ success rate for the DoorKey domain with distractor objects.}
		\label{table:gridworld}
\end{table}

The results in Table~\ref{table:gridworld} (averaged over 10 trials) show that \emph{LgTS} reaches a successful policy quicker compared to the LFS, TSCL, and LgRS. The overall number of interactions to learn a set of successful policies for satisfying the high-level goal objective are higher in presence of distractor objects because of low level agent interactions with these objects and the increased dimensionality of the state space of the RL agent. 
For the experiment with distractor objects, we observed a graph edit distance of $3.4 \pm 0.4$ between the LgTS approach and the graph generated using the AgTS approach, which is higher than the graph edit distance that was computed without the presence of distractor objects. This difference indicates that the graphs generated using the LgTS approach did contain paths that involved distractor objects, however, the graph also contained paths that did not involve the distractor object and the RL agent was able to learn successful policies for such paths.

\subsection{LgTS - modified prompt}

Recent approaches that use LLM for task guidance have a curated prompt and a fine-tuned LLM that prevents generalization to newer prompts that have similar meaning but different descriptors. This finetuning prevents generalization to unseen out-of-distribution prompts and descriptors.
To demonstrate how the prompt influences the LLM output which in turn affects learning progress, we evaluated LgTS by changing the prompt to the LLM. In this test, a fraction (at random) of entity and predicate descriptors were changed to a synonym chosen from Thesaurus~\cite{dictionary2002merriam} (for e.g., Key was replaced with Code; Door was replaced with Gate).

\begin{table}
    		\centering
	\begin{tabular}{>{\centering}p{1.1cm}>{\centering}p{2.6cm}>{\centering}p{2.4cm}p{0.05cm}}
\arrayrulecolor{taupegray}\toprule
\textbf{Approach} & \textbf{$\#$ Interactions} \\ (Mean $\pm$ SD) & \textbf{Success Rate }\\ (Mean $\pm$ SD) \tabularnewline
\midrule
\rowcolor{gainsboro}
\emph{LgTS}  & $(4.87    \pm 0.74)\times 10^6$ &\centering $0.81 \pm 0.09$\tabularnewline
\rowcolor{gainsboro}
\emph{AgTS}  & $(2.67 \pm 0.36)\times 10^6$ &\centering $0.94 \pm 0.02$\tabularnewline
\emph{LFS}  &  $5 \times 10^7$ &\centering $0 \pm 0$\tabularnewline
AgRS  & $5 \times 10^7$ &\centering $0.05 \pm 0.04$ \tabularnewline
LgRS  & $5 \times 10^7$ &\centering $0 \pm 0$ \tabularnewline
TSCL  & $5 \times 10^7$ &\centering $0 \pm 0$ \tabularnewline
\bottomrule
\end{tabular}
		\caption{Table comparing $\#$interactions $\&$ success rate for the DoorKey domain with modified prompt.}
		\label{table:gridworld}
\end{table}
The results in Table~\ref{table:gridworld} (averaged over 10 trials) show that \emph{LgTS} reaches a successful policy quicker compared to the LFS, TSCL, and LgRS. The overall number of interactions to learn a set of policies that satisfy the high-level goal objective are higher when the prompt was changed as compared to LgTS with a constant and curated prompt. We observed that the LLM was able to accommodate the new prompt and suggest paths that satisfied the high-level objective.

\subsection{LgTS - Search and Rescue task }

To demonstrate how LgTS performs when the plan length becomes deeper, we evaluated LgTS on a more complex urban Search and Rescue domain. In this domain, the agent acts in a grid setting where it needs to perform a series of sequential sub-tasks to accomplish the final goal of the task. The agent needs to open a door using a key, then collect a fire extinguisher to extinguish the fire, and then find and rescue stranded survivors. A fully-connected graph generated using the above mentioned high-level states consists of 24 distinct transitions. This is a multi-goal task as the agent needs to extinguish fire as well as rescue survivors to reach the goal state.
We use the LLM to produce seven distinct high-level paths that help prune transitions that the LLM does not recommend while providing little information about the environment as possible.  


\begin{table}
    		\centering
	\begin{tabular}{>{\centering}p{1.1cm}>{\centering}p{2.6cm}>{\centering}p{2.4cm}p{0.05cm}}
\arrayrulecolor{taupegray}\toprule
\textbf{Approach} & \textbf{$\#$ Interactions} \\ (Mean $\pm$ SD) & \textbf{Success Rate }\\ (Mean $\pm$ SD) \tabularnewline
\midrule
\rowcolor{gainsboro}
\emph{LgTS}  & $(1.13 \pm 0.26)\times 10^7$ &\centering $0.76 \pm 0.11$\tabularnewline
\rowcolor{gainsboro}
\emph{AgTS}  & $(8.61 \pm 0.12)\times 10^6$ &\centering $0.87 \pm 0.04$\tabularnewline
\emph{LFS}  &  $5 \times 10^6$ &\centering $0 \pm 0$\tabularnewline
AgRS  & $5 \times 10^7$ &\centering $0.05 \pm 0.04$ \tabularnewline
LgRS  & $5 \times 10^7$ &\centering $0 \pm 0$ \tabularnewline
TSCL  & $5 \times 10^7$ &\centering $0 \pm 0$ \tabularnewline
\bottomrule
\end{tabular}
		\caption{Table comparing $\#$interactions $\&$ success rate for the Search and Rescue domain.}
		\label{table:gridworld}
\end{table}

The results in Table~\ref{table:gridworld} (averaged over 10 trials) show that \emph{LgTS} reaches a successful policy quicker compared to the LFS, TSCL, and LgRS. The overall number of interactions to learn a set of successful policies for satisfying the high-level goal objective are higher when the prompt was changed as compared to LgTS with a constant and curated prompt. We observed that the LLM was able to accommodate the new prompt and suggest paths that satisfied the high-level goal objective.

\subsection{Discussion}
We designed a method that queries an LLM to produce sub-goal sequences based on entities and predicates known about the task.  Each entity and predicate is assumed to have a single word natural language description. An off-the-shelf LLM is prone to associate certain entities with certain predicates based on its training data and procedure. For e.g., when we attempted to make the search and rescue task even more complex by adding a \textit{debris} element that needs to be moved using the \textit{moving} predicate, we observed that the LLM is associating the predicate with other entities already present in the environment, such as fire extinguisher, door etc. Since the LLM does not have access to the type hierarchy that associates predicates with entities, the LLM is conflicted when two similar entities are applicable to the same predicate. As an experiment, we also provided the type hierarchies to the LLM and we observed that the graph generated using LgTS had a graph edit distance of $4.6$ compared to the graph given by an oracle, which was lower than the graph edit distance observed without the presence of type hierarchies, which was found to be $7.3$. Thus, incorporating information that informs the LLM about predicate-entity associations helps produce sub-goal sequences that are semantically closer to the sub-goals given by the LTL$_f$ formula suggested by an oracle. 

The prompts generated by the LLM also depend on the type of language model used. When we changed our LLM from LLAMA2 to GPT-4 on the complex search and rescue task mentioned above, we observed a graph edit distance of $5.1$ compared to the graph given by an oracle, which was lower than the GED for LLAMA2, which was $7.3$. This shows that GPT-4 was successful in producing responses and in turn graphs which were closer to the graph generated from an oracle. With further advancements in the LLM capabilities, we might observe even further improvements in the reasoning ability of such models, which in turn will produce better and meaningful entity-predicate associations. Existing tools such as chain-of-thought prompting and access to a dataset that can finetune the LLM to produce valid and useful outputs will further improve the prediction accuracy of the LLM. However, even with such advancements, the environmental configuration will be unknown to an agent that does not have access to the transition dynamics model. This work is a step in that direction. LgTS attempts to bridge the gap between the LLM-generated sub-goal outputs and the policies that an agent can learn to satisfy these sub-goals while minimizing the number of times it interacts with the environment.  
\section{Conclusion and Future Work}
We proposed LgTS, a framework for dynamic task sampling
for RL agents using a graphical representation of sub-goal sequences suggested by a large language model. Through experiments, we demonstrated that LgTS accelerates learning,
converging to a desired success rate quicker as compared to
other curriculum learning baselines and achieves comparable success compared to sub-goal sequences provided by an oracle. We also evaluated our approach on a complex long-horizon search and rescue task where the number of predicates and entities were higher and the agent needed to satisfy several sub-goals to satisfy the final goal objective. LgTS reduced training time
without relying on human-guided dense reward function. LgTS accelerates learning when information about the entities present in the environment and the sensors that can identify the truth assignment of predicates is available.\\
\\
\\
\textbf{Limitations \& Future Work:} In certain cases, the natural language description of the entities and the predicates might be incorrect or unavailable. In that case, the sub-goal sequences suggested by the LLM will be based on these incorrect descriptions, and the sequences might harm the learning progress of the agent. 
Our future plans involve automating the entity identification process that will eliminate the need to rely on predefined entities. In case of robotic environments, this can be done using an object detector along with a pose estimator that can identify the natural language description of objects in the environment along with relative position. 
Our approach recognizes and discards sub-tasks for which policies exist that can satisfy the sub-tasks' goal objective. This minimizes the number of interactions with the environment by avoiding policy learning for a number of sub-tasks.
As an extension, we would like to explore biasing away from sub-tasks rather than completely discarding them once the target node is
reached, so in the limit, optimal or near-optimal policies can be obtained. We
would also like to explore complex robotic and multi-agent
scenarios with elaborate goal directed objectives. On the LLM front, we would like to incorporate closed-loop feedback from the RL agent to the LLM that promotes improved response generation by the LLM.





\bibliography{main}


\end{document}